\documentclass[lettersize,journal]{IEEEtran}
\usepackage{amsmath,amsfonts}
\usepackage{algorithmic}
\usepackage{algorithm}
\usepackage{array}
\usepackage[caption=false,font=normalsize,labelfont=sf,textfont=sf]{subfig}
\usepackage{textcomp}
\usepackage{stfloats}
\usepackage{url}
\usepackage{verbatim}
\usepackage{graphicx}
\usepackage{cite}
\usepackage{microtype}      
\usepackage{xcolor}         
\usepackage{graphicx}
\usepackage{amsmath}
\usepackage{multicol}
\usepackage{multirow}
\usepackage{natbib}
\usepackage{amssymb}%
\usepackage{pifont}%
\usepackage{booktabs}       
\usepackage{amsfonts}       
\usepackage{nicefrac}      
\usepackage{url}      
\newcommand{\cmark}{\ding{51}}%
\newcommand{\xmark}{\ding{55}}%
\usepackage{hyperref}

\hypersetup{urlcolor=blue}

\begin{document}

\title{HResFormer: Hybrid Residual Transformer for Volumetric Medical Image Segmentation}

\author{Sucheng Ren and Xiaomeng Li \thanks{S. Ren and X. Li are with Department of Electronic and Computer
		Engineering, The Hong Kong University of Science and Technology, Hong Kong SAR, China. X. Li is also with HKUST Shenzhen-Hong Kong Collaborative Innovation Research Institute, Futian, Shenzhen, China.  This work was partially supported by the National Natural Science Foundation of China under Grant 62306254, a grant from the Hong Kong Innovation and Technology Fund for Projects PRP/041/22FX, the Project of Hetao Shenzhen Hong Kong Science and Technology Innovation Cooperation Zone (HZQB-KCZYB-2020083), and a grant from the Research Grants Council of the Hong Kong Special Administrative Region, China (Project Reference Number: T45-401/22-N). Corresponding author: Xiaomeng Li.}}

\markboth{Journal of \LaTeX\ Class Files,~Vol.~14, No.~8, August~2021}%
{Shell \MakeLowercase{\textit{et al.}}: A Sample Article Using IEEEtran.cls for IEEE Journals}

\maketitle

\begin{abstract}
Vision Transformer shows great superiority in medical image segmentation due to the ability in learning long-range dependency. For medical image segmentation from 3D data, such as computed tomography (CT), existing methods can be broadly classified into 2D-based and 3D-based methods. One key limitation in 2D-based methods is that the intra-slice information is ignored, while the limitation in 3D-based methods is the high computation cost and memory consumption, resulting in a limited feature representation for inner-slice information. During the clinical examination, radiologists primarily use the axial plane and then routinely review both axial and coronal planes to form a 3D understanding of anatomy. Motivated by this fact, our key insight is to design a hybrid model which can first learn fine-grained inner-slice information and then generate a 3D understanding of anatomy by incorporating 3D information. We present a novel \textbf{H}ybrid \textbf{Res}idual trans\textbf{Former} \textbf{(HResFormer)} for 3D medical image segmentation. Building upon standard 2D and 3D Transformer backbones, HResFormer involves two novel key designs: \textbf{(1)} a \textbf{H}ybrid \textbf{L}ocal-\textbf{G}lobal fusion \textbf{M}odule \textbf{(HLGM)} to effectively and adaptively fuse inner-slice information from 2D Transformer and intra-slice information from 3D volumes for 3D Transformer with local fine-grained and global long-range representation. \textbf{(2)} a residual learning of the hybrid model, which can effectively leverage the inner-slice and intra-slice information for better 3D understanding of anatomy. 
Experiments show that our HResFormer outperforms prior art on widely-used medical image segmentation benchmarks. This paper sheds light on an important but neglected way to design Transformers for 3D medical image segmentation. 
\end{abstract}


\section{Introduction}

Vision Transformer (ViT)~\citep{dosovitskiy2020image,liu2021swin,wang2021pyramid} is a powerful tool in the field of computer vision, dominating tasks such as classification, detection, and segmentation. One key factor in its success is the self-attention core module~\citep{vaswani2017attention,valanarasu2021medical}, which allows for learning long-range dependencies and overcomes the limitations of locality found in traditional convolution.
Recently, many efforts have been focused on adapting Transformers for use in medical imaging segmentation from 3D scans such as computed tomography (CT) and magnetic resonance imaging (MRI). 
These methods can be broadly divided into two categories: 2D-based methods~\citep{transunet,valanarasu2021medical,gao2021utnet,swinunet} and 3D-based methods~\citep{nnformer,cotr,unetr}. 
In the first category, the 3D volume is sliced into 2D images. The result is obtained by first feeding them into various network architectures, then concatenating the 2D predictions into a 3D prediction.  
For example, TransUnet~\citep{transunet} integrated the ViT with a U-shape convolution network for medical image segmentation. SwinUnet~\citep{swinunet} introduced hierarchical Swin Transformer~\citep{liu2021swin} into a U-shape encoder and decoder for multi-organ and cardiac segmentation.
For the second category, recent works introduced 3D Transformer~\citep{cotr,nnformer,unetr} for medical image segmentation. For example, CoTR~\citep{cotr} incorporated 3D CNN and 3D  deformable Transformer to extract local and long-range features for accurate 3D medical image segmentation. 
UNTER~\citep{unetr} reformed the 3D volumes into sequences and captured the global information across the whole input volumes.

Although these methods~\citep{cotr,swinunet,transunet,unetr,nnformer,nnunet} showed promising results, they have the following limitations: (1) 2D-based methods learn features based on a single slice instead of a 3D volume, which ignores the intra-slice information and lacks the 3D understanding of the whole anatomy. 
(2) Existing 3D-based methods require high computation and memory consumption since the whole 3D volume is equally cut into sequences, and computation cost is quadratic to the length of sequences. 
Such a design learns results in granularity representations for axial view, the primary plane used by radiologists for interpretation, leading to a limited feature representation. 

During the examination, radiologists primarily use the axial plane and then routinely review both axial and coronal planes to form a 3D understanding of anatomy. Motivated by this fact, our key insight is to design a hybrid model which can first learn fine-grained inner-slice information and then generate a 3D understanding of anatomy by incorporating 3D volumetric information. To this end, we present a novel \textbf{H}ybrid \textbf{R}esidual trans\textbf{Former} \textbf{(HResFormer)} for 3D medical image segmentation.
It builds upon a standard 2D network to learn fine-grained inner-slice features. To effectively learn 3D information, a \textbf{H}ybrid \textbf{L}ocal-\textbf{G}lobal fusion \textbf{M}odule \textbf{(HLGM)} is proposed, which takes the 2D predictions as prior knowledge and mutually fuse 3D information from 3D data with fine-grained local and global long-range information for the 3D Transformer. We further propose the residual learning of a hybrid network to facilitate the 3D understanding of anatomical structure. 
This paper has three contributions:
\begin{itemize}
	\item Unlike prior art that directly designs 2D or 3D-based operations, this is the first hybrid Transformer, shedding light on an important but largely neglected way for 3D medical image segmentation.
	\item We propose a novel HResFormer and it also involves two key technical innovations: an HLGM to effectively fuse inner-slice information to 3D Transformer and a residual learning to facilitate the 3D understanding of anatomy.

	\item Our method consistently previous outperforms state-of-the-art methods on widely-used medical image segmentation benchmarks. Code will be available at \href{https://github.com/xmed-lab/HResFormer}{https://github.com/xmed-lab/HResFormer}. 
\end{itemize}

\section{Related Work}

\textbf{CNN-based Segmentation Methods.}
In the deep learning era, convolution-based methods show great improvements over traditional hand-crafted features based methods in medical image segmentation~\citep{ronneberger2015u,cciccek20163d,nnunet,li20183d}. FANet~\citep{tnnls1} proposes feedback attention for 2D biomedical imaging datasets and SwinPA-Net~\citep{tnnls2} proposes a dense multiplicative connection (DMC) module and local pyramid attention for polyp segmentation task and skin lesion segmentation task, while our HResFormer designs a hybrid model that can first learn fine-grained inner-slice information (2D) and then generate 3D understanding of anatomy by incorporating 3D information. \citep{tnnls3} designs a novel Dynamic Loss and evaluates it on multiorgan segmentation which is included in our paper. However, our paper focus on architecture design which is orthogonal to~\citep{tnnls3}. DS-TransUnet ~\citep{tnnls4}, GRENet ~\citep{tnnls5} and EC-CaM-UNet ~\citep{tnnls6} design 2D models for 2D medical image segmentation.

Particularly, U-shape models with skip connection~\citep{ronneberger2015u,li2018h,nnunet} dominate the medical image analysis and show tremendous success for various medical image segmentation tasks.  \citep{li2021more} developed a 2.5D 24-layer Fully Convolutional Network with boundary loss for liver and tumor segmentation tasks where the residual block was incorporated into the model. As most medical data are 3D volumes like CT, MRI, the 3D convolution neural network is used for mining information cross three spatial dimensions simultaneously. 3D UNet~\citep{cciccek20163d} directly extend the 2D convolution in UNet operation to 3D. However, the heavy computation costs of 3D convolution limit the depth and width of the 3D UNet model. Res-UNet~\citep{xiao2018weighted} and 3DRUNet~\citep{lee2017superhuman} introduce residual connections into 3D Unet to avoid gradient vanishing. LAMP~\citep{zhu2020lamp} introduces the automated model parallelism to train large 3D models for efficient liver segmentation. \citep{kim2019scalable} propose to search convolution neural network architecture for 3D medical image segmentation by continuous relaxation stochastic sampling strategies.  Among these methods,  nnUnet~\citep{nnunet}, one of the representative convolution-based models, shows the state-of-the-art performance on multiple medical image segmentation benchmarks.

\textbf{Transformer-based Segmentation Methods.} Researchers introduced various Transformer-based methods for 3D medical image segmentation. The first line of research is building 2D-based networks~\citep{transunet, zhang2021transfuse}, in which the 3D volumes are sliced into 2D slices and the results are the stack of outputs. For example, TransUnet~\citep{transunet} incorporated ViT with a U-shape network for medical image segmentation. TransFuse~\citep{zhang2021transfuse} proposed a BiFusion module to fuse features from both convolution-based and Transformer-based networks.
SwinUnet~\citep{swinunet} proposed a pure Unet-style Swin~\citep{liu2021swin} Transformer-based encoder-decoder for medical image segmentation. DS-TransUNet~\citep{lin2022ds} extend SwinUnet to multiscale paradigm with an extra encoder and a fusion module to interact with multscale features globally. \citep{li2021more} introduce Transformer into decoder instead of encoder to upsample. \citep{youclass} incorporate Transformer and Generative Adversarial Network for medical segmentation. MCTrans~\citep{ji2021multi} combines complex feature learning and rich semantic structure excavating into a unified framework by Multi-Compound Transformer by mapping the convolutional features from different multi-scale to different tokens and performs intra- and inter-scale self-attention, rather than single-scale attention. RTNet\citep{huang2022rtnet} design relation self-attention and build the transformer with two capabilities including global dependencies among lesion features, while a cross-attention transformer
allows interactions between lesion and vessel features by
integrating valuable vascular information to alleviate ambiguity in lesion detection caused by complex fundus structures.

The second line of research is building 3D-based networks~\citep{wang2021transbts, cotr, nnformer, unetr}, in which the 3D volumes are taken as the input. For example, \citep{karimi2021convolution} build a convolution-free model by split 3D volume into 3D patches for mapping 3D volumes into 1D sequences and feed into Transformer.
TransBTS~\citep{wang2021transbts} integrated the Transformer into a 3D encoder-decoder network for brain tumor segmentation from 3D MR images. To address the computation cost of Transformer in processing high-resolution images, CoTR~\citep{cotr} incorporated 3D CNN and 3D deformable Transformer to learn effective features. UTNER~\citep{unetr} reformulated the task of 3D medical image segmentation as a sequence-to-sequence prediction problem. 
 BAT~\citep{wang2021boundary} introduces modeling long-range dependency and local fine-grained details simultaneously by boundary-aware Transformer. T-AutoML~\citep{yang2021t} searches both best architecture and hyperparameter with Transformer architecture for  Lesion Segmentation. SpecTr~\citep{yun2021spectr} introduces Transformer by formulating contextual feature learning across spectral bands for hyperspectral pathology image segmentation as a sequence-to-sequence processing procedure. Swin UNETR ~\citep{hatamizadeh2022swin} maps multimodal data into 1D sequence and feeds them into a hierarchical Swin transformer which provides features at different levels and the shift-window attention keeps the abilities of global preservation. nnFormer~\citep{nnformer} introduced a volume-based self-attention instead of 2D self-attention to learn 3D representations. SwinBTS~\citep{Jiang2022SwinBTSAM} share a similar framework with SwinUnet but extend it to 3D version with an enhanced transformer module for detailed feature extraction. PHTrans~\citep{liu2022phtrans} propose a hybrid architecture for medical image segmentation which parallelly hybridizes Transformer and CNN in main building blocks to produce hierarchical representations from global and local features and adaptively aggregate them.
Unlike prior 2D or 3D-based methods, our HResFormer is the first hybrid Transformer for 3D medical image segmentation, which effectively learns fine-grained inner-slice information and intra-slice information, contributing to a better 3D understanding of anatomy. 
 
\section{Method}

\begin{figure*}[t]
\centering
\begin{tabular}{c}
\includegraphics[scale=0.48]{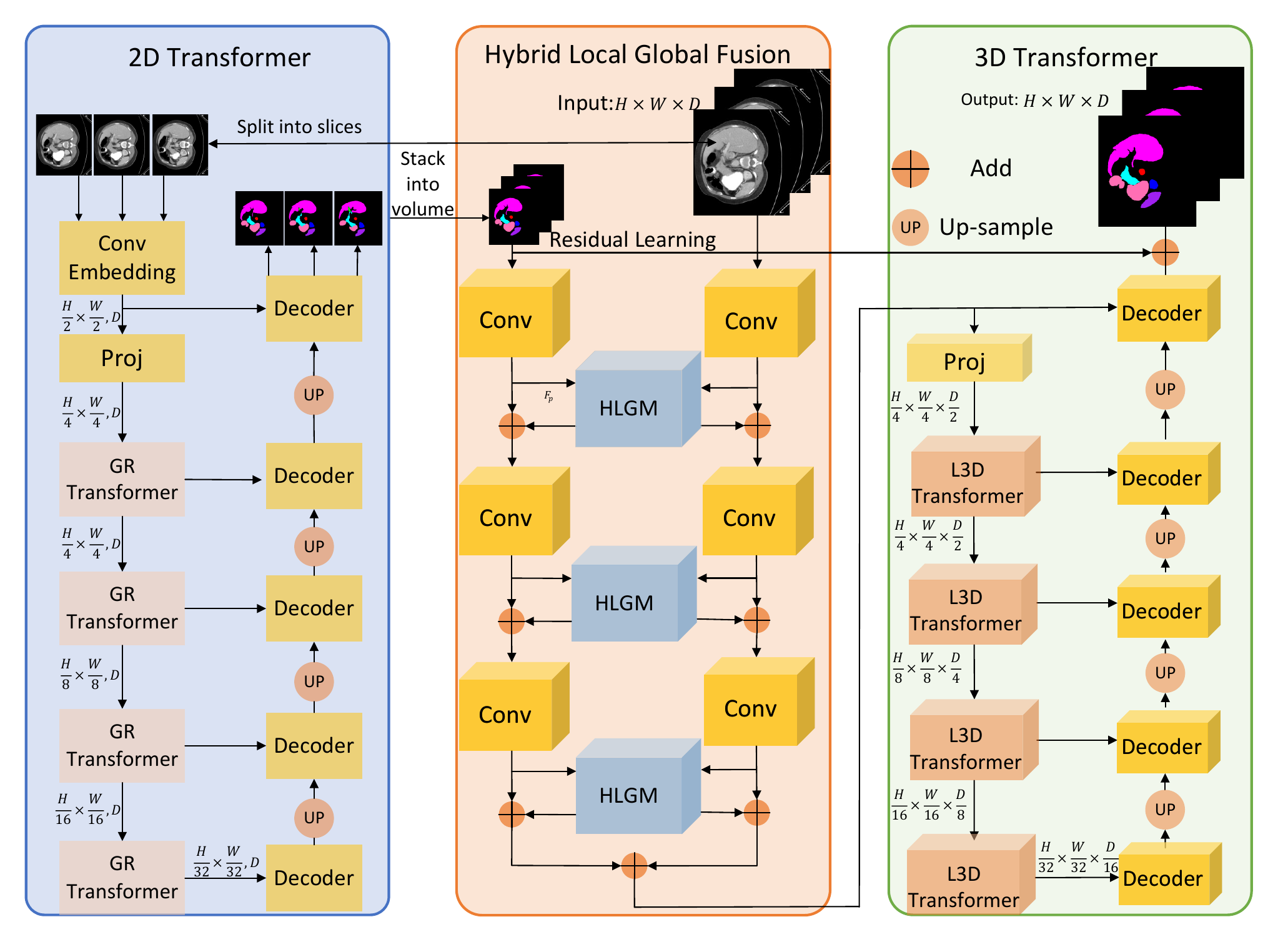}
\end{tabular}
\caption{Overview of our HResFormer. It consists of three main parts: a 2D model, a Hybrid Local-Global fusion Module (HLGM), and a 3D model. The 3D volume is split into slices, and the 2D model makes predictions slice by slice to generate fine-grained inner-slice information. Then, the 2D predictions are stacked into a volume (we still call them as 2D predictions) and are fused with the 3D volume locally and globally via HLGM. Finally, the 3D model takes the fused features as input and incorporates the 2D predictions with residual learning for the better understanding of 3D anatomy with both inner-slice and inter-slice representations.}
\label{img:method}
\end{figure*}

Figure~\ref{img:method} shows the overview of our HResFormer, which consists of three main parts: a 2D-based Transformer to learn fine-grained inner-slice information, a Hybrid Local-Global fusion Module (HLGM) for effectively fusing 2D predictions and 3D data with both local fine-grained and global representations, and a 3D-based Transformer for the better 3D understanding of anatomy with inner-slice information as a prior.

\subsection{2D-based Transformer Backbone}
We first slice the 3D volume into a stack of 2D slices and feed them into a 2D-based Transformer. By concatenating the 2D predictions, we can obtain the 3D output for each input volume. 
Our 2D model is an encoder-decoder architecture with a Transformer-based encoder and super-light weight convolution (only very few convolution-layer norm-GeLU layers) based decoder. The Transformer-based encoder contains a patch embedding layer and a few Transformer blocks. 
For the patch embedding layer, inspired by~\citep{wang2021scaled}, we take Convolutional Patch Embedding to replace naive patch embedding which simply splits the image into non-overlap patches. Before feeding $F_{2d,0}$ into the following Transformer blocks, we take a projection layer to adjust the size and channel of the feature maps. 
Compared to existing methods~\citep{transunet,unetr} that use a  single scale ViT based-Transformer cross all blocks, we take the Transformer with pyramid architecture as the encoder, which provides feature maps with different resolutions for decoder and enables the effective feature learning for segmentation. In practice, we split the model into 4 stages with the feature maps size of $4\times, 8\times, 16\times, 32\times$ downsampling compared with the input size instead of a single resolution of $16\times$ downsampling in ViT. Specifically, in the $i_{th}$ stage, there are $L_i$ Transformer blocks with the feature map size of $F_{2d,i} \in \mathcal{R}^{\frac{H}{2^{i+1}} \times \frac{W}{2^{i+1}}\times  C^i}$. With these five features, the light-weight decoder can  decode the features to generate high-resolution predictions.  

\textbf{Convolutional Patch Embedding.}  Previous Transformers~\citep{touvron2021training,dosovitskiy2020image} simply take one large stride convolution to map the image into sequence and reduce the size of the feature map 16$\times$ compared to the input size. However, recent researches~\citep{wang2021scaled} find that taking several successive convolution layers with overlaps to replace the naive non-overlap patch embedding can provide a better input sequence for the Transformer and boost the performance of Transformer. Based on this consideration, we take two convolution-layer norm-GeLU layers to replace the non-overlap patch embedding.

\textbf{Global Reduction (GR) Transformer Blocks.}
Each Global Reduction Transformer block~(GR Transformer) contains a Global Reduction Multi-head Self-Attention
module~(GR-MSA), followed by a feed-forward layer (FFD) including 2-layer MLP with GELU activation. Besides, we apply the LayerNorm (LN) layer before each multi-head self-attention module and MLP layer. A residual connection is applied after each module.

The pyramid architecture of our 2D model requires the self-attention layer to apply on large-size feature maps like $\frac{H}{4} \times \frac{W}{4}$, while ViT applies self-attention on small-size feature maps $\frac{H}{16} \times \frac{W}{16}$. However, the original self-attention layer is quadratic to the size of feature maps, resulting in a high computation cost and memory consumption. 
Previous work~\citep{swinunet,liu2021swin,nnformer} takes the window attention from Swin Transformer and limits the self-attention to a local region. Such window attentions limit the ability of self-attention to perceive the whole feature maps. In contrast, we aim to preserve the ability to capture global dependency.

Inspired by Pyramid Vision Transformer~\citep{wang2021pyramid,ren2022shunted,j1,j2,r2,r3}, we merge tokens to reduce the computation cost with Global Reduction Attention (GR-Attention) with different scale. Give the feature $\mathbf{F}$, we calculate the Query, Key, Value and reduce the spatial size of K and V:
\begin{equation}
\begin{split}
	Q &= W_q*F ,\\
	K &= W_k*\text{GR}(F, r),\\
	V &= W_v*\text{GR}(F, r),
	\end{split}
\end{equation}
where $\text{GR}(F, r)$ indicates down-sampling the spatial size $r$ times. Different $r$ help self-attention layer capture different scale information. In practice, we implement $\text{GR}(F, r)$ by a convolution layer with stride $r$. Our Global Reduction Attention ($\text{GR-MSA}$) significantly reduce the computation cost with the Query, and length reduced Key and Value compared with vanilla self-attention.

\begin{align}
	\begin{split}
		F_{\text{GR}}^{l}&=\text{GR-MSA}\left(\text{Norm}\left(F_{\text{2d}}^{l-1}\right)\right)+F_{\text{2d}}^{l-1},\\
		F_{\text{2d}}^{l}&=\text{MLP}\left(\text{Norm}\left(F_{\text{GR}}^{l}\right)\right)+F_{\text{GR}}^{l},
	\end{split}
\end{align}
At the same time, there is a depth-wise convolution branch parallel to Global Reduction Attention. Except for the attention layer, the Transformer block contains a Feed-Forward layer with two MLP layers. However, such a Feed-Forward layer is positional-invariant, and there is only cross channel information but no cross position information in the Feed-Forward layer. Therefore, we propose to take Cross Position Feed Forward Layer seen in Sec. \ref{LGF} to replace the traditional Feed-Forward layer to complement the cross position information.

With the four features from the encoder, the decoder decodes them all for the 2D predictions with skip connection. During the training, our 2D model outputs the predictions $P_{2d, i}$ for deep supervision~\citep{lee2015deeply} at each stage. 

\subsection{3D-based Transformer Backbone}
The 2D predictions provide the inner slice's fine-grained predictions while lacking inter-slice information. Given the predictions from the 2D model as prior knowledge, the 3D counterpart can effectively learn 3D volumetric features.  %

\textbf{Patch Embedding.} Unlike previous 3D models~\citep{unetr,nnunet,nnformer} that use 3D volume as the input, our 3D model takes both 3D volume and 2D predictions generated by 2D Transformer as the input. Unlike the widely used patch embedding, to better integrate the predictions and 3D input volumes, we propose a Hybrid Local-Global fusion module in Sec.~\ref{LGF} to replace the vanilla patch embedding.

\textbf{Local 3D Transformer Blocks.} We pass the features from the Hybrid Local-Global fusion module into 3D Transformer blocks. Inspired by the Swin and video Swin Transformer~\citep{liu2021video,liu2021swin}, we propose to split the input features into non-overlap 3D regions and perform the self-attention within each region. Give the feature $F_{3d}$, we split it into $N = \frac{H}{wd} \times \frac{W}{wd} \times \frac{D}{wd}$ non-overlap region. The local 3D self-attention (L3D-MSA) performs locally in each region and in parallel across different regions. 
To build cross-region information interaction, we take shift local 3D attention (SL3D-MSA) the same as Swin Transformer to shift the region partition, which changes the connection between different positions in consecutive blocks. To sum up, the Local 3D Transformer (L3D Transformer) is:
\begin{align}
	\begin{split}
		\hat{F}_{\text{3D}}^{l}&=\text{L3D-MSA}\left(\text{Norm}\left(F_{\text{3D}}^{l-1}\right)\right)+F_{\text{3D}}^{l-1},\\
		F_{\text{3D}}^{l}&=\text{MLP}\left(\text{Norm}\left(\hat{F}_{\text{3D}}^{l}\right)\right)+\hat{F}_{\text{3D}}^{l},\\
		\hat{F}_{\text{3D}}^{l+1}&=\text{SL3D-MSA}\left(\text{Norm}\left(F_{\text{3D}}^{l}\right)\right)+F_{\text{3D}}^{l},\\
		F_{\text{3D}}^{l+1}&=\text{MLP}\left(\text{Norm}\left(\hat{F}_{\text{3D}}^{l+1}\right)\right)+\hat{F}_{\text{3D}}^{l+1}.\\
	\end{split}
\end{align}
At the same time, there is a 3D depth-wise convolution branch parallel to local 3D self-attention. Similar to the 2D model, except for the attention layer, the Transformer block contains a Feed-Forward layer with two MLP layers. Therefore, we propose to take the Cross Position Feed Forward Layer seen in Sec. \ref{LGF} to replace the traditional Feed-Forward layer.

\subsection{Hybrid Local-Global Fusion Module (HLGM)}
\begin{figure}[t]
    \centering
    \includegraphics[width=\linewidth]{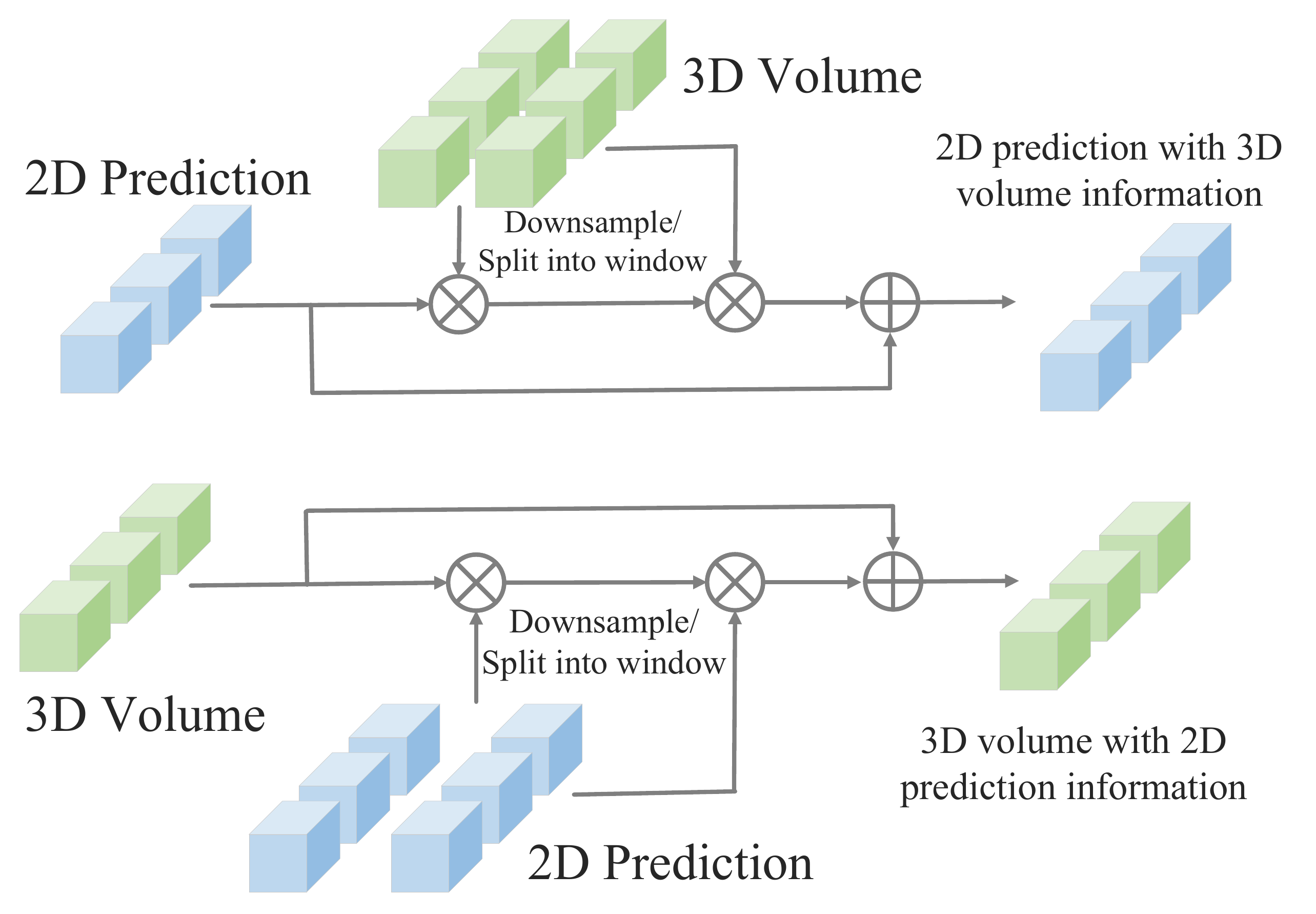}
    \caption{The overview of Hybrid Local-Global Fusion Module. We use one module to complement 2D prediction features to 3D volume features and 3D volume features to 2D prediction features.}
    \label{fig:hlgfm}
\end{figure}
\label{LGF}
As shown in Figure \ref{fig:hlgfm}, to facilitate the 3D understanding of anatomy, we propose a Hybrid Local-Global fusion Module (HLGM) to selectively and adaptively fuse 2D predictions $P_{2d}$ and 3D input volume $X$. %
Our HLGM consists of a local mutual fusion (LMF) group, a global mutual fusion (GMF) group and a cross position feed forward (CPFF) layer. 
\begin{equation}
\begin{split}
    &\hat{F}_{p}^{i} = \text{LMF}(F_{p}^{i-1}, F_{v}^{i-1})\oplus\text{GMF}(F_{p}^{i-1}, F_{v}^{i-1}) + F_{p}^{i-1}, \\
    &F_{p}^{i} = \text{CPFF}(\hat{F}_{p}^{i}) + \hat{F}_{p}^{i},\\
    &\hat{F}_{v}^{i} = \text{LMF}(F_{v}^{i-1}, F_{p}^{i-1})\oplus\text{GMF}(F_{v}^{i-1}, F_{p}^{i-1}) + F_{v}^{i-1}, \\
    &F_{v}^{i} = \text{CPFF}(\hat{F}_{v}^{i}) + \hat{F}_{v}^{i},\\
\end{split}
\end{equation}
where the features $F_{p}$ are from 2D predictions and $F_{v}$ are from 3D data. We map the $F_{p}$ and $F_{v}$ into local groups and global groups to capture local details and global semantics, respectively.

\textbf{Local Mutual Fusion (LMF) Group.}
The goal of LMF is to locally and mutually fuse fine-grained information from 2D predictions and 3D volume. Formally, we divide features into non-overlap volumes N = $S_H \times S_W \times S_D$ and apply local mutual fusion $\text{LMF}(F_{p}, F_{
v})$ to fuse features of 3D volumes. Note that compared to window attention in~\citep{liu2021swin}, we have two differences: 1) we fuse these two features in a 3D local region instead of a 2D local window. 2) we mutually model cross-feature fusion dependency between features from 2D prediction and 3D volumes instead of self-feature dependency. 
\begin{equation}
    \begin{split}
    &q_{p} = W_q*(F_{p}),  \\
    &k_{v} = W_k*(F_{v}), \\
    &v_{v} = W_v*(F_{v}), \\
    &\hat{F}_{p, local} =  \text{L-MSA}(q_{p}, k_{v}, v_{v}) 
    \end{split}
\end{equation}
where $\text{L-MSA}$ indicate perform the self-attention in the local region. Similarly, we apply local mutual fusion $\text{LMF}(F_{v}, F_{p})$ to fuse features of 2D prediction:
\begin{equation}
    \begin{split}
    &q_{v}= \text{MLP}(F_{v}), \\
    &k_{p}= \text{MLP}(F_{p}), \\
    & v_{p}= \text{MLP}(F_{p}) \\
    &\hat{F}_{v,local} =  \text{L-MSA}(q_{v}, k_{p}, v_{p}) 
    \end{split}
\end{equation}
To replenishment the global information, we also propose Global Mutual Fusion.

\begin{table*}[t]
    \begin{center}
    	\caption{Results of multi-organ segmentation on the Synapse dataset with 8 classes. The evaluation metric is DSC (\%). Best results are bolded. $^\bigtriangledown$ denotes using a larger input size (512$\times$512). ``T'' refers to Transformer-based and ``C'' refers to convolution-based methods. The results are adopted from nnFormer~\citep{nnformer}.}
	\resizebox{0.98\textwidth}{!}{
	\begin{tabular}{l|c|c|cccccccc} 
	\toprule
	    \multirow{2}{*}{Methods}    & \multirow{2}{*}{Type} & Average & \multirow{2}{*}{Ao} & \multirow{2}{*}{Gb}  & \multirow{2}{*}{Ki(L)}  & \multirow{2}{*}{Ki(R)}  & \multirow{2}{*}{Li}  & \multirow{2}{*}{Pa} & \multirow{2}{*}{Sp}  & \multirow{2}{*}{St} \\
	    &    & DSC $\uparrow$ &&&&&&&&\\
		\hline
		\hline
		ViT \citep{dosovitskiy2020image} + CUP \citep{transunet} & 2D, T& 67.86 & 70.19 & 45.10 & 74.70 & 67.40 & 91.32 & 42.00 & 81.75 & 70.44\\
		R50-ViT \citep{dosovitskiy2020image} + CUP \citep{transunet} &  2D, T& 71.29 & 73.73 & 55.13 & 75.80 & 72.20 & 91.51 & 45.99 & 81.99 & 73.95\\
        TransUNet~\citep{transunet}&	 2D, T& 77.48 &  87.23 & 63.16 & 81.87 & 77.02 & 94.08 & 55.86 & 85.08 & 75.62\\
        TransUNet{$^\bigtriangledown$}~\citep{transunet}	&  2D, T& 84.36 & 90.68 & 71.99 & 86.04 & 83.71 & 95.54 & 73.96 & 88.80 & 84.20\\
        SwinUNet~\citep{swinunet} & 2D, T& 79.13 & 85.47 & 66.53 &83.28 &79.61 & 94.29 & 56.58 & 90.66 & 76.60\\
        TransClaw U-Net~\citep{chang2021transclaw}  &  2D, T & 78.09 & 85.87 & 61.38 & 84.83 & 79.36 & 94.28 & 57.65 & 87.74 &73.55\\
        LeVit-UNet-384s~\citep{xu2021levit}&  2D, T&78.53 &87.33&62.23	&84.61	&80.25	&93.11	&59.07	&88.86	&72.76\\
        MISSFormer~\citep{huang2021missformer}&  2D, T& 81.96 & 86.99 & 68.65 & 85.21 & 82.00 & 94.41 & 65.67 & 91.92 & 80.81\\
        \hline 
        V-Net~\citep{milletari2016v} &  3D, C&68.81 & 75.34& 51.87& 77.10& 80.75& 87.84& 40.05& 80.56& 56.98 \\
        UNETR~\citep{unetr}&  3D, T& 79.56 & 89.99 & 60.56 & 85.66 & 84.80 & 94.46 & 59.25 & 87.81 & 73.99 \\
        nnFormer~\citep{nnformer}&  3D, T& 86.57 & 92.04 & 70.17 &86.57 & 86.25 & 96.84 & 83.35 & 90.51 & 86.83\\
        nnUnet~\citep{nnunet}&   3D, C& 86.99  &93.01 &  71.77  & 85.57 & 88.18  &\textbf{ 97.23}& 83.01  & 91.86& 85.26\\
        HResFormer (\textbf{Ours})&   Hybrid, T & \textbf{89.10} & \textbf{93.83} & 71.90 & 90.37 & 91.57 & 96.84 & \textbf{83.68} & \textbf{96.40} & \textbf{88.15}\\
        \bottomrule
	\end{tabular}}
	\label{tab:synapse8}
	\end{center}
\end{table*}

\begin{table*}[ht]
    \begin{center}
    	\caption{Results of multi-organ segmentation on the Synapse dataset with 13 organ classes. }
	\resizebox{0.98\textwidth}{!}{
	\begin{tabular}{l|c|ccccccccccccc} 
	\toprule
	    \multirow{2}{*}{Methods}  & Average & \multirow{2}{*}{Sp} & \multirow{2}{*}{Ki(R)}  & \multirow{2}{*}{Ki(L)}  & \multirow{2}{*}{Gb}  & \multirow{2}{*}{Es}  & \multirow{2}{*}{Li} & \multirow{2}{*}{St}  & \multirow{2}{*}{Ao}& \multirow{2}{*}{ICV} & \multirow{2}{*}{Ve} & \multirow{2}{*}{Pa} & \multirow{2}{*}{Ad(R)} & \multirow{2}{*}{Ad(L)}  \\
	    &  DSC $\uparrow$ &&&&&&&&\\
	    \hline
	    \hline 
        CoTR~\citep{cotr} & 85.00& 95.77&  94.24& 92.92& 76.39& 76.61& 97.03&86.65&91.06 & 87.31&  77.69& 82.54& 72.32& 74.44\\
        nnUnet~\citep{nnunet}& 85.58& 96.64&  94.74& 92.54& 76.33& 76.58& 97.05&88.15&90.95 & 88.53&  77.82& 83.87& 73.47& 75.88\\
        \hline
        HResFormer (\textbf{Ours}) & \textbf{87.46} & 96.40&  94.58& 93.37& 81.90& 79.39& 96.85&89.15&92.84 & 91.42&  83.39& 84.69& 75.41& 79.55\\
        \bottomrule
	\end{tabular}}

	\label{tab:synapse13}
	\end{center}
\end{table*}

\textbf{Global Mutual Fusion (GMF) Group.}
The goal of GMF is to globally fuse long-range information learned from 2D predictions and the 3D volumes.  
Note that there are two differences between our GMF and the self-attention in pyramid vision Transformer~\citep{wang2021pyramid}: 1) we fuse these two features in a 3D and reduce spatial dimension in all three dimensions. 2) we mutually model cross-feature fusion dependency instead of self-feature dependency between features from 2D prediction and 3D volume. We apply global mutual fusion GMF$(F_{p}, F_{v})$ to fuse features from 2D prediction.
\begin{equation}
    \begin{split}
    &q_{p}= W_p*F_{p}, \\
    &k_{v} = W_k*\text{SR}(F_{v}, r),\\
    &v_{v} = W_v*\text{SR}(F_{v}, r)  \\
    &\hat{F}_{p, global} =  \text{MSA}(q_{p}, k_{v}, v_{v}) 
    \end{split}
\end{equation}
Where $\text{SR}$ is spatial reduction to reduce the length of Key and Value implemented by 3D convolution. Similarly, we apply global mutual fusion GMF$(F_{v}, F_{p})$ to fuse features from 2D prediction:
\begin{equation}
    \begin{split}
    &q_{v}= W_q*(F_{v}), \\
    &k_{p}= W_k*\text{SR}(F_{p}, r), \\
    &v_{p}=  W_v*\text{SR}(F_{p}, r) \\
    &\hat{F}_{v, global} = \text{MSA}(q_{v}, k_{p}, v_{p}) 
    \end{split}
\end{equation}
With global and local mutual fusion, slices by slices prediction and 3D data selectively and adaptive fused in local fine-granularity and global coarse-granularity.

\textbf{Cross Position Feed Forward (CPFF) Layer}
Traditional Feed Forward layer positional independently process features, therefore, we aim at complementing cross position details in the feed forward layer.
\begin{equation}
    \begin{split}
        &F^{'} = \text{MLP}(F)\\
        &F^{''} = \text{MLP}(\phi(F^{'}+ \text{CP}(F^{'})))\\
        &\hat{F} = \phi(F^{''}+\text{CP}(F^{''}))\\
    \end{split}
\end{equation}
where $\phi$ is GeLU activation layer.
In practice, we simply take depth-wise convolution as the cross position layer ($CP$).

\subsection{Residual Learning of the Hybrid Model}
Previous work~\citep{nnformer,nnunet,cotr} training 3D model with segmentation target which segments the data and predict different sub-region directly. In contrast, we propose to take training 3D model as residual learning to complement inter-slices information for 2D prediction:
\begin{equation}
    P^{'}_{3d, i} = P_{2d, 0} + P_{3d, i}
\end{equation}
where $P^{'}_{3d}$ indicates the final prediction of our HResFormer, $S_{2d}$ and $P_{3d}$ indicate the output of 2D and 3D model, respectively. 
To train the HResFormer, we take both cross entropy loss ($l_{ce}$)~\citep{ce} and dice loss($l_{d}$)~\citep{dice} as our loss function on 4 outputs from 2D model $P_{2d}$ and 4 outputs from 3D model with residual prediction of 2D model $P^{'}_{3d}$. During inference, we only take $P^{'}_{3d,0}$ as the final output.

\section{Experiments}

\subsection{Datasets}
\textbf{Synapse dataset for multi-organ segmentation~\footnote{https://www.synapse.org/Synapse:syn3193805/wiki/217789}}
This dataset contains 30 abdominal CT scan cases with 13 organs. All the labels are annotated by the interpreters under supervision of clinical radiologists at Vanderbilt University Medical Center. We follow two settings: 1) the settings of nnFormer~\citep{nnformer}: 18 cases used for training and 12 cases used for testing with 8 class segmentation  2) the settings of nnUnet~\citep{nnunet}: 21 cases used for training and 9 for testing with 13 organ segmentation. We take the average Dice-Similarity coefficient (DSC) as an evaluation metric. 

\textbf{Brain Tumor MRI scans Segmentation~\citep{bakas2018identifying}} This dataset contains 484 multimodal MRI data, including four channels: FLAIR, T1w, T1gd, and T2w. There are three sub-regions: Edema (ED), Non-enhancing tumor(NET), and Enhancing tumor (ET). We follow the same setting of~\citep{nnformer} to split data into training, validation and test dataset.

\textbf{Automated Cardiac Diagnosis (ACDC)~\citep{bernard2018deep}.}  ACDC is the MRI scans collected from 100 different patients. There are three labels for each case: left ventricle (LV), right ventricle (RV) and myocardium (MYO). We following the same settings of nnformer, and take 70 samples as training set, 10 samples as validation set, and 20 samples as test set.

\subsection{Implementation Details}
The parameter is initialized by the ImageNet pre-trained model if it exists, and the rest Transformer blocks are randomly initialized, following the default settings of DeiT~\citep{touvron2021training} and PyTorch. We take SGD with momentum 0.9 as the optimizer. The learning rate starts from 0.01 and gradually reduces with “poly” decay strategy. The total training epoch is 1000. We follow the same prepossessing strategies~\citep{nnunet,nnformer} including rotation, scaling, Gaussian noise, Gaussian blur, brightness and contrast adjust, simulation of low resolution, gamma augmentation, and mirroring.

\begin{table*}[t]
\begin{center}
 \caption{
 Brain tumor segmentation results on the multimodal MRI dataset. %
 Best results are bolded. Experimental results of baselines are adopted from~\citep{unetr}. }
    	\resizebox{0.6\textwidth}{!}{\begin{tabular}{l|c|ccc}
    \toprule
       \multirow{2}{*}{Methods}  & Average & \multirow{2}{*}{Edema}& Non-Enhancing  & Enhancing \\ 
       &  DSC $\uparrow$  & &Tumor & Tumor \\
       \hline
       \hline
       SETR NUP~\citep{zheng2021rethinking} & 63.7 & 69.7 & 54.4 &  66.9 \\
       SETR PUP~\citep{zheng2021rethinking} & 63.8  & 69.6  & 54.9 &  67.0 \\
       SETR MLA~\citep{zheng2021rethinking} & 63.9  & 69.8  & 55.4 & 66.5 \\
       TransUNet~\citep{transunet} &  64.4 &  70.6 & 54.2 &  68.4 \\
       TransBTS~\citep{wang2021transbts}  & 69.6 &  77.9 &57.4 & 73.5 \\
       CoTr w/o CNN~\citep{cotr} & 64.4  & 71.2& 52.3 &  69.8 \\
       CoTr~\citep{cotr} &  68.3  & 74.6  & 55.7  & 74.8 \\
       UNETR~\citep{unetr}  & 71.1  & 78.9  & 58.5 & 76.1 \\
       \hline
       nnFormer~\citep{nnformer} &77.4& 84.4  &66.7 & 81.0\\
       nnUnet~\citep{nnunet}  & 77.7 & 83.8  & 67.3 & 81.8\\
       
       HResFormer (\textbf{Ours}) & \textbf{79.5} & \textbf{86.2}  & \textbf{69.1} & \textbf{83.3}\\
    \bottomrule
    \end{tabular}}
    \label{tb_bts}
\end{center}
\end{table*}

\begin{table}[htbp]
    \begin{center}
    \small
	\caption{Comparison of model parameters for Transformer-based methods.}
	\begin{tabular}{l|ccccc} 
	\toprule
	    Method& TransUnet& UNETR&nnFormer&HResFormer\\
		\midrule
		Parameters &105M &102M&150M&117M\\
        \bottomrule
	\end{tabular}
	\label{tab:param}
	\end{center}
\end{table}

\begin{table}[]
\normalsize
    \centering
    \caption{Computation cost comparison with nnFormer and HResFormer. HResFormer outperforms nnFormer with less computation cost.}
    \begin{tabular}{l|c|c}
    \toprule
        Method &Flops (G) & Synapse(DSC) \\
        \midrule
        nnFormer &   157.9& 86.6\\
        HResFormer -lightweight&63.4&87.2 \\
        HResFormer  &131.7 &89.1\\
    \bottomrule
    \end{tabular}
    
    \label{tab:flops}
\end{table}

\begin{table*}[htbp]
    \begin{center}
    	\caption{Effectiveness of each component of our HResFormer on the Synapse dataset. }
	\resizebox{0.98\textwidth}{!}{
	\begin{tabular}{l|cccc|c|cccccccc} 
	\toprule
	    \multirow{2}{*}{Methods}  &\multicolumn{4}{c|}{Configuration}  & Average & \multirow{2}{*}{Ao} & \multirow{2}{*}{Gb}  & \multirow{2}{*}{Ki(L)}  & \multirow{2}{*}{Ki(R)}  & \multirow{2}{*}{Li}  & \multirow{2}{*}{Pa} & \multirow{2}{*}{Sp}  & \multirow{2}{*}{St} \\
	    & 2D&3D&HGLM&Residual &  DSC $\uparrow$ &&&&&&&&\\
		\hline
		\hline
         MISSFormer~\citep{huang2021missformer}&\cmark& \xmark& \xmark&	\xmark & 81.96 & 86.99 & 68.65 & 85.21 & 82.00 & 94.41 & 65.67 & 91.92 & 80.81\\ 
        HResFormer (only 2D)	&\cmark& \xmark & \xmark & \xmark& 81.05 &  85.31 & 67.07 & 85.67 & 80.11 & 94.44 & 68.38 & 87.46 & 79.98\\
        \hline
        nnFormer~\citep{nnformer}& \xmark  &\cmark& \xmark & \xmark &  86.57 & 92.04 & 70.17 &86.57 & 86.25 & 96.84 & 83.35 & 90.51 & 86.83\\
        nnUnet~\citep{nnunet}& \xmark &\cmark& \xmark & \xmark & 86.99  &93.01 &  71.77  & 85.57 & 88.18  &\textbf{ 97.23}& 83.01  & 91.86& 85.26\\
        HResFormer (only 3D)  & \xmark&\cmark& \xmark& \xmark& 85.39 & 92.04 & 70.17 &86.57 & 86.25 & 96.84 & 83.35 & 90.51 & 86.83\\
        \hline
        \multirow{4}{*}{HResFormer} &\cmark&\cmark&\xmark& \xmark & 87.23 & 92.27 & 70.11 &87.58 & 89.12 & 95.65 & 82.44 & 93.67 & 87.03\\
        &\cmark&\cmark&\cmark& \xmark & 88.21 & 93.04 & 70.39 &88.97 & 90.96 & 96.14 & 83.35 & 94.81 & 88.03\\
        &\cmark&\cmark& \xmark & \cmark &  87.81 & 92.34 & 70.99 & 88.59 & 89.53 & 96.58 & 83.11 & 94.15 &87.25\\
        &\cmark&\cmark&\cmark& \cmark&  \textbf{89.10} & \textbf{93.83} & \textbf{71.90} & \textbf{90.37} & \textbf{91.57}  & 96.84 & \textbf{83.68} & \textbf{96.40} & \textbf{88.15}\\
        \bottomrule
	\end{tabular}}
	\label{tab:hybrid}
	\end{center}
\end{table*}

\begin{figure*}[h]
	\centering
	\setlength{\tabcolsep}{.5pt}
	\renewcommand{\arraystretch}{.5}
	\begin{tabular}{ccccc}
		\includegraphics[width=.18\linewidth]{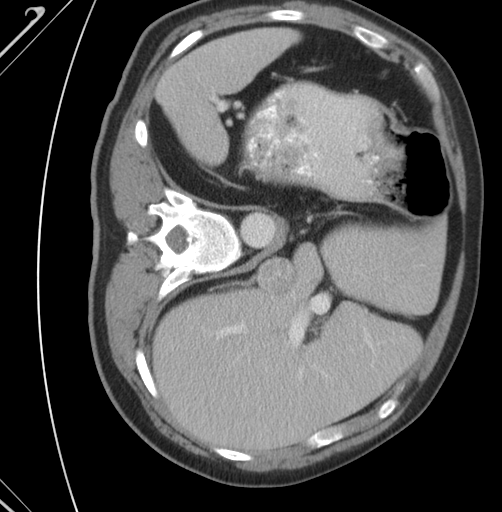}&
		\includegraphics[width=.18\linewidth]{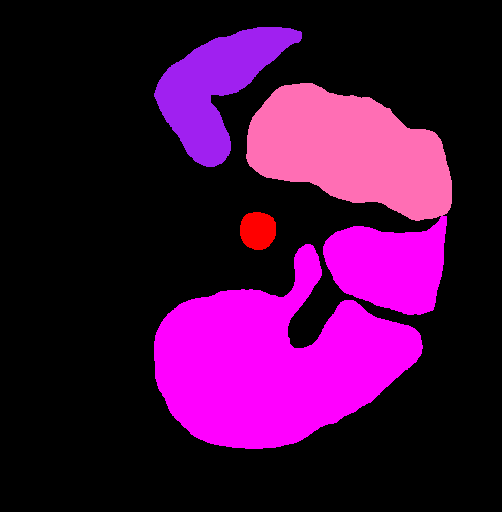}&
		\includegraphics[width=.18\linewidth]{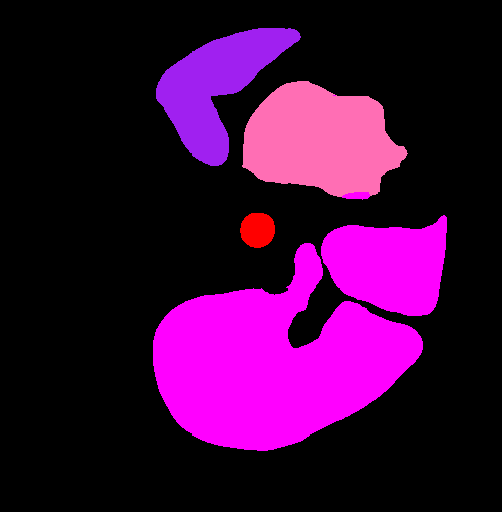}&
		\includegraphics[width=.18\linewidth]{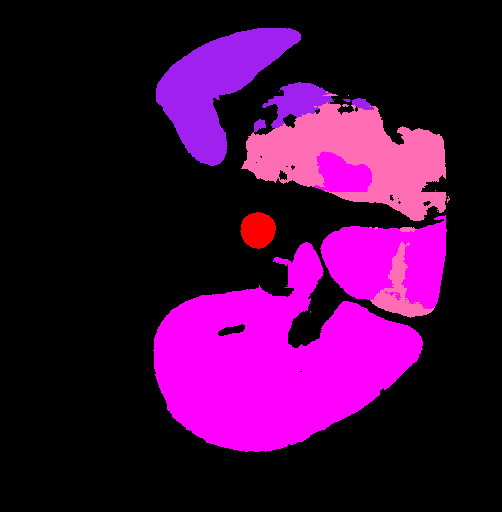}&
		\includegraphics[width=.18\linewidth]{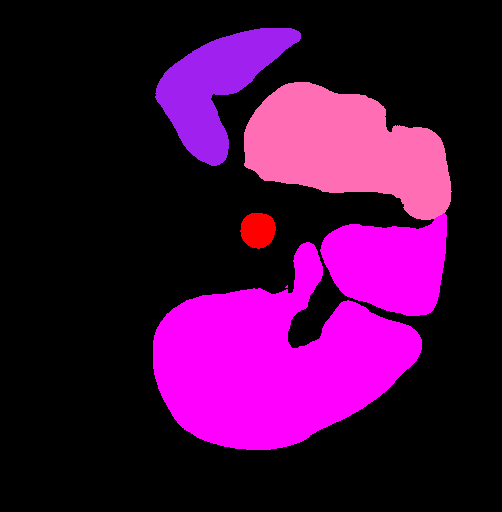}\\

		Input & GT & 2D & 3D  & HResFormer (Ours)
	\end{tabular}
	\caption{Examples of multi-organ segmentation results.  ``GT'' refers to the ground-truth. \textcolor[rgb]{1.,0,0}{Red}, \textcolor[rgb]{1.,0,1.}{magenta}, \textcolor[rgb]{1,0.43137,0.705882}{pink}, \textcolor[rgb]{0.62745,0.12549,0.9412}{purple} refer to aotra (Ao), liver (Li), stomach (St) and spleen (Sp), respectively. ``2D'' or ``3D'' refers to HResFormer (only 2D) and HResFormer (only 3D) in Table~\ref{tab:hybrid}.}
	\label{img:23h}
\end{figure*}

\begin{table*}[h]
    \begin{center}
    	\caption{Ablation study of our HLGM. We compare with alternative fusion methods on the Synapse dataset. ``Na\"{\i}ve'' refers to stacking 2D predictions into a volume and concatenating it with the 3D raw data. ``FF'' refers to performing the ``Na\"{\i}ve'' on the feature level. ``DFF'' refers to fusing deep features from convolution layers that share the identical structure with our HResFormer, but without our HLGM. }
\resizebox{0.8\textwidth}{!}{	\begin{tabular}{l|c|c|c|c|c|c|c|c|c} 
	\toprule
	    \multirow{2}{*}{Methods}  & Average & \multirow{2}{*}{Ao} & \multirow{2}{*}{Gb}  & \multirow{2}{*}{Ki(L)}  & \multirow{2}{*}{Ki(R)}  & \multirow{2}{*}{Li}  & \multirow{2}{*}{Pa} & \multirow{2}{*}{Sp}  & \multirow{2}{*}{St} \\
	    &   DSC $\uparrow$ &&&&&&&&\\
		\hline
		\hline
        Ours-Na\"{\i}ve &  87.81 & 92.34 & 70.99 & 88.59 & 89.53 & 96.58 & 83.11 & 94.15 &87.25\\
        Ours-FF & 87.87& 92.15 & 71.01 & 88.68 & 89.57 & 96.54 & 83.18 & 94.05 & 87.15\\
        Ours-DFF & 88.18 & 92.61 & 70.91 & 89.72 & 90.21 & 96.86 & 82.88 & 95.08&87.19\\
        Ours-HLGM  & \textbf{89.10} & 93.83 & 71.90 & 90.37& 91.57 & 96.84 & 83.68 & 96.40 & 88.15\\
        \bottomrule
	\end{tabular}}
	\label{tab:fusion}
	\end{center}
\end{table*}

\begin{table}[htbp]
    \centering
    \caption{Results of segmentation on the ACDC dataset. }
    \resizebox{0.48\textwidth}{!}{
    \begin{tabular}{c|c|ccc}
\hline
Methods    & DSC            & RV             & MLV            & LVC            \\ \hline
Swin-Une~\citep{swinunet}  & 90.00          & 88.55          & 85.62          & \textbf{95.83} \\
TransUNet\citep{transunet}  & 89.71          & 88.86          & 84.53          & 95.73          \\
LeViT-Unet~\citep{xu2021levit} & 90.32          & 89.55          & 87.64          & 93.76          \\
MISSFormer~\citep{huang2021missformer} & 90.86          & 89.55          & 88.04          & 94.99          \\
nnFormer~\citep{nnformer}   & 91.62          & \textbf{90.27} & 89.23          & 95.36          \\
nnU-Net~\citep{nnunet}    & 91.36          & 90.11          & 88.75          & 95.23          \\ \hline

Ours    & \textbf{91.88} & 90.44         & \textbf{89.50} & 95.79          \\ \hline
\end{tabular}}
    
    \label{tab:acdc}
\end{table}

\subsection{Comparison with other State-of-the-art Methods}
\textbf{Results on Synapse multi-organ CT dataset.} As shown in Table~\ref{tab:synapse8} and Table~\ref{tab:synapse13}, we compare our method with state-of-the-art methods under two widely-used settings. 
For the settings of nnFormer in Table~\ref{tab:synapse8}, 2D model like TransUNet~\citep{transunet} and SwinUnet~\citep{swinunet} which segment 3D volumes slices by slices achieves similar performance with simple 3D model UNETR~\citep{unetr} and V-net~\citep{milletari2016v}, which shows the effectiveness of 2D models.
However, with the inner slice information a well-designed 3D model like nnUnet and nnFormer outperforms the 2D model by about 6\% on DSC. Our HRFomrer unifies the advantages of both 2D and 3D models, achieves the best performance under two settings, and outperforms previous Transformer-based state-of-the-art nnFormer~\citep{nnformer} by 2.5\% on DSC and convolution-based state-of-the-art nnUnet~\citep{nnunet} by 2.1\% on DSC, respectively. 
Table~\ref{tab:synapse13} shows our HResFormer outperforms CoTR~\citep{cotr} by 2.4\% on DSC and nnUnet~\citep{nnunet} by 1.8\% on DSC, respectively. Notably, our HResFormer consistently outperforms nnUnet on all 13 organs. Noted that with such promising performance gain, our HResFormer has similar parameters to Transformer-based methods~\citep{nnformer,transunet,unetr} seen in Table~\ref{tab:param} and lower computation cost to nnFormer. Besides, we also report a lightweight version of HResFormer which has half the computation cost and still outperforms nnFormer.

\textbf{Results on Multimodal MRI dataset.} 
We report the brain tumor segmentation results in Table~\ref{tb_bts}.  It is clear that our HResFormer consistently outperforms previous methods in all three sub-regions: Edema (ED), Non-enhancing tumor (NET), and Enhancing tumor (ET). To be specific, our HResFormer outperforms previous convolution-based state-of-the-art nnUnet~\citep{nnunet} by 1.8\% on DSC and Transformer-based state-of-the-art nnFormer~\citep{nnformer} by 2.1\% on DSC, respectively.

\textbf{Results on Automated Cardiac Diagnosis (ACDC).} As shown in Table \ref{tab:acdc}, 2D Transformers like SwinUnet, TransUnet achieve high performance with over 90 DSC which shows the effectiveness of 2D model. While the model goes to 3D, the performance make further improvements. Our Hybrid model take both advantages of 2D and 3D Transformers and achieves the best performance. To be specific, our HResFormer outperforms nnUnet by 0.52 on DSC.

\subsection{Ablation Study of our HResFormer}
In ablation study, all the experiments are conducted on the Synapse dataset, following the same setting of nnFormer~\citep{nnformer}.

\textbf{Effectiveness of each component of our HResFormer.} Table~\ref{tab:hybrid} shows the effectiveness of our HResFormer from the following aspects: \textbf{(1)  Pure 2D-based and 3D-based network}. Compared to the state-of-the-arts, we can see that our 2D model and 3D model achieve slightly lower results with 81.05\% vs. 81.96\%~\citep{huang2021missformer} for 2D and 85.39\% vs. 86.57\%~\citep{nnformer} for 3D. 
The comparison shows that the effectiveness of our HResFormer is not due to the strong backbones in pure 2D and 3D models.  
\textbf{(2) Effectiveness of HGLM.}
Table~\ref{tab:hybrid} shows that our HGLM can enhance the model performance from 87.23\% to 88.21\%, demonstrating the effective design of HGLM in fusing the mutual information between 2D and 3D network. 
\textbf{(3) Effectiveness of residual learning.} 
With residual learning, we can see that the performance consistently improves from 88.21\% and 89.10\%, showing that residual learning can further enhance the segmentation results generated by the 3D model. 
The design of residual learning of HResFormer can effectively leverage the fine-grained inner-slice information learned by the 2D network and intra-slice information generated by the 3D network. Figure~\ref{img:23h} visualizes the segmentation results of the 2D and 3D parts of HResFormer, where we can see that our HResFormer can effectively enhance the performance.

\textbf{Ablation study of HLGM.}
To prove the effectiveness of our HLGM in fusing local and global information from 2D and 3D network, we compare it with three kinds of baseline fusion methods: 
\emph{Na\"{\i}ve Fusion}, \emph{Feature Fusion.} and \emph{Deep Feature Fusion}. 
\textbf{\emph{Na\"{\i}ve Fusion (Ours-Na\"{\i}ve)}} refers to stacking the 2D predictions into a volume and concatenating it with the 3D volume as the input to the 3D model.
\textbf{\emph{Feature Fusion (Ours-FF)}} refers to performing the  \emph{Na\"{\i}ve Fusion (Ours-Na\"{\i}ve)} on the feature level to densely fuse the inner-slice knowledge learned from 2D network and intra-slice knowledge learned from 3D network. 
Recent research find the sequence from several consecutive convolution layers is better for the following Transformer blocks.
\textbf{\emph{Deep Feature Fusion (Ours-DFF)}} refers to fusing the deep features from several convolution layers which share the identical structure with our HResFormer, but without the HLGM.

We can observe from Table~\ref{tab:fusion} that a Na\"{\i}ve fusion method can achieve 87.81\% on DSC and outperforms previous state-of-the-arts, showing that the hybrid model is one the of the important but neglected solutions to 3D medical image segmentation. However, simply performing the Na\"{\i}ve fusion on the feature level (Ours-FF) and fusing deep features from convolution layers (Ours-DFF) cannot significantly improve the performance. 
Compared to the one without HLGM (Ours-DFF), we can see that our HLGM can improve the overall segmentation result by 1.0\% on DSC over 8 classes. 
The comparison shows that HLGM can effectively and adaptively fuse local (inner-slice and fine-grained) and global (intra-slice and long-range) information between 2D and 3D networks.

\section{Conclusion}
\label{conclusion}
This paper presents a novel hybrid residual Transformer for 3D medical image segmentation, where the key insight is to mimic the radiologists: first learn fine-grained inner-slice features from the axial plane and then form a 3D understanding with intra-slice information. HResFormer excels other state-of-the-art methods on two widely-used 3D medical image segmentation benchmarks.
\paragraph{Limitations.} The performance on test images from unknown distributions is not guaranteed since we follow the standard supervised learning protocols. Expanding training datasets to include a wider variety of organ shapes, sizes, and pathological conditions can improve the model's generalization capabilities.
\paragraph{Future work.} The core component of our HResFormer is the attention mechanism, which exhibits quadratic computational complexity relative to sequence length. Recent advancements, such as Mamba~\citep{mamba1,mamba2}, introduce modules that achieve a global receptive field with linear complexity in language models. We are considering integrating these modules to replace our current attention mechanism, aiming for further enhancements in computational efficiency. Besides, it is worth to explore HResFormer working together with recent foundation model like SAM~\citep{sam}.

\bibliographystyle{IEEEtranN}
\small{\bibliography{refs}}

\begin{thebibliography}{61}
\providecommand{\natexlab}[1]{#1}
\providecommand{\url}[1]{#1}
\csname url@samestyle\endcsname
\providecommand{\newblock}{\relax}
\providecommand{\bibinfo}[2]{#2}
\providecommand{\BIBentrySTDinterwordspacing}{\spaceskip=0pt\relax}
\providecommand{\BIBentryALTinterwordstretchfactor}{4}
\providecommand{\BIBentryALTinterwordspacing}{\spaceskip=\fontdimen2\font plus
\BIBentryALTinterwordstretchfactor\fontdimen3\font minus \fontdimen4\font\relax}
\providecommand{\BIBforeignlanguage}[2]{{%
\expandafter\ifx\csname l@#1\endcsname\relax
\typeout{** WARNING: IEEEtranN.bst: No hyphenation pattern has been}%
\typeout{** loaded for the language `#1'. Using the pattern for}%
\typeout{** the default language instead.}%
\else
\language=\csname l@#1\endcsname
\fi
#2}}
\providecommand{\BIBdecl}{\relax}
\BIBdecl

\bibitem[Dosovitskiy et~al.(2021)Dosovitskiy, Beyer, Kolesnikov, Weissenborn, Zhai, Unterthiner, Dehghani, Minderer, Heigold, Gelly, et~al.]{dosovitskiy2020image}
A.~Dosovitskiy, L.~Beyer, A.~Kolesnikov, D.~Weissenborn, X.~Zhai, T.~Unterthiner, M.~Dehghani, M.~Minderer, G.~Heigold, S.~Gelly \emph{et~al.}, ``An image is worth 16x16 words: Transformers for image recognition at scale,'' in \emph{ICLR}, 2021.

\bibitem[Liu et~al.(2021)Liu, Lin, Cao, Hu, Wei, Zhang, Lin, and Guo]{liu2021swin}
Z.~Liu, Y.~Lin, Y.~Cao, H.~Hu, Y.~Wei, Z.~Zhang, S.~Lin, and B.~Guo, ``Swin transformer: Hierarchical vision transformer using shifted windows,'' in \emph{ICCV}, 2021, pp. 10\,012--10\,022.

\bibitem[Wang et~al.(2021{\natexlab{a}})Wang, Xie, Li, Fan, Song, Liang, Lu, Luo, and Shao]{wang2021pyramid}
W.~Wang, E.~Xie, X.~Li, D.-P. Fan, K.~Song, D.~Liang, T.~Lu, P.~Luo, and L.~Shao, ``Pyramid vision transformer: A versatile backbone for dense prediction without convolutions,'' in \emph{ICCV}, 2021, pp. 568--578.

\bibitem[Vaswani et~al.(2017)Vaswani, Shazeer, Parmar, Uszkoreit, Jones, Gomez, et~al.]{vaswani2017attention}
A.~Vaswani, N.~Shazeer, N.~Parmar, J.~Uszkoreit, L.~Jones, A.~N. Gomez \emph{et~al.}, ``Attention is all you need,'' in \emph{NeurIPS}, 2017, pp. 5998--6008.

\bibitem[Valanarasu et~al.(2021)Valanarasu, Oza, Hacihaliloglu, and Patel]{valanarasu2021medical}
J.~M.~J. Valanarasu, P.~Oza, I.~Hacihaliloglu, and V.~M. Patel, ``Medical transformer: Gated axial-attention for medical image segmentation,'' in \emph{MICCAI}.\hskip 1em plus 0.5em minus 0.4em\relax Springer, 2021, pp. 36--46.

\bibitem[Chen et~al.(2021)Chen, Lu, Yu, Luo, Adeli, Wang, et~al.]{transunet}
J.~Chen, Y.~Lu, Q.~Yu, X.~Luo, E.~Adeli, Y.~Wang \emph{et~al.}, ``{TransUNet}: Transformers make strong encoders for medical image segmentation,'' \emph{arXiv preprint arXiv:2102.04306}, 2021.

\bibitem[Gao et~al.(2021)Gao, Zhou, and Metaxas]{gao2021utnet}
Y.~Gao, M.~Zhou, and D.~N. Metaxas, ``Utnet: a hybrid transformer architecture for medical image segmentation,'' in \emph{MICCAI}.\hskip 1em plus 0.5em minus 0.4em\relax Springer, 2021, pp. 61--71.

\bibitem[Cao et~al.(2022)Cao, Wang, Chen, Jiang, Zhang, Tian, and Wang]{swinunet}
H.~Cao, Y.~Wang, J.~Chen, D.~Jiang, X.~Zhang, Q.~Tian, and M.~Wang, ``Swin-unet: Unet-like pure transformer for medical image segmentation,'' in \emph{ECCV Workshops}, 2022.

\bibitem[Zhou et~al.(2023)Zhou, Guo, Zhang, Han, Yu, Wang, and Yu]{nnformer}
H.-Y. Zhou, J.~Guo, Y.~Zhang, X.~Han, L.~Yu, L.~Wang, and Y.~Yu, ``nnformer: Volumetric medical image segmentation via a 3d transformer,'' \emph{IEEE Transactions on Image Processing}, vol.~32, pp. 4036--4045, 2023.

\bibitem[Xie et~al.(2021)Xie, Zhang, Shen, and Xia]{cotr}
Y.~Xie, J.~Zhang, C.~Shen, and Y.~Xia, ``Cotr: Efficiently bridging cnn and transformer for 3d medical image segmentation,'' in \emph{MICCAI}.\hskip 1em plus 0.5em minus 0.4em\relax Springer, 2021, pp. 171--180.

\bibitem[Hatamizadeh et~al.(2022{\natexlab{a}})Hatamizadeh, Tang, Nath, Yang, Myronenko, Landman, Roth, and Xu]{unetr}
A.~Hatamizadeh, Y.~Tang, V.~Nath, D.~Yang, A.~Myronenko, B.~Landman, H.~R. Roth, and D.~Xu, ``Unetr: Transformers for 3d medical image segmentation,'' in \emph{WACV}, 2022, pp. 574--584.

\bibitem[Isensee et~al.(2021)Isensee, Jaeger, Kohl, Petersen, and Maier-Hein]{nnunet}
F.~Isensee, P.~F. Jaeger, S.~A. Kohl, J.~Petersen, and K.~H. Maier-Hein, ``nnu-net: a self-configuring method for deep learning-based biomedical image segmentation,'' \emph{Nature methods}, vol.~18, no.~2, pp. 203--211, 2021.

\bibitem[Ronneberger et~al.(2015)Ronneberger, Fischer, and Brox]{ronneberger2015u}
O.~Ronneberger, P.~Fischer, and T.~Brox, ``{U-Net}: Convolutional networks for biomedical image segmentation,'' in \emph{MICCAI}.\hskip 1em plus 0.5em minus 0.4em\relax Springer, 2015, pp. 234--241.

\bibitem[{\c{C}}i{\c{c}}ek et~al.(2016){\c{C}}i{\c{c}}ek, Abdulkadir, Lienkamp, Brox, and Ronneberger]{cciccek20163d}
{\"O}.~{\c{C}}i{\c{c}}ek, A.~Abdulkadir, S.~S. Lienkamp, T.~Brox, and O.~Ronneberger, ``3d u-net: learning dense volumetric segmentation from sparse annotation,'' in \emph{MICCAI}.\hskip 1em plus 0.5em minus 0.4em\relax Springer, 2016, pp. 424--432.

\bibitem[Li et~al.(2018{\natexlab{a}})Li, Dou, Chen, Fu, Qi, Belav{\`y}, Armbrecht, Felsenberg, Zheng, and Heng]{li20183d}
X.~Li, Q.~Dou, H.~Chen, C.-W. Fu, X.~Qi, D.~L. Belav{\`y}, G.~Armbrecht, D.~Felsenberg, G.~Zheng, and P.-A. Heng, ``3d multi-scale fcn with random modality voxel dropout learning for intervertebral disc localization and segmentation from multi-modality mr images,'' \emph{Medical image analysis}, vol.~45, pp. 41--54, 2018.

\bibitem[Tomar et~al.(2022)Tomar, Jha, Riegler, Johansen, Johansen, Rittscher, Halvorsen, and Ali]{tnnls1}
N.~K. Tomar, D.~Jha, M.~A. Riegler, H.~D. Johansen, D.~Johansen, J.~Rittscher, P.~Halvorsen, and S.~Ali, ``Fanet: A feedback attention network for improved biomedical image segmentation,'' \emph{IEEE Transactions on Neural Networks and Learning Systems}, 2022.

\bibitem[Du et~al.(2022)Du, Wang, Liu, Wang, and Meijering]{tnnls2}
H.~Du, J.~Wang, M.~Liu, Y.~Wang, and E.~Meijering, ``Swinpa-net: Swin transformer-based multiscale feature pyramid aggregation network for medical image segmentation,'' \emph{IEEE Transactions on Neural Networks and Learning Systems}, 2022.

\bibitem[Song et~al.(2023{\natexlab{a}})Song, Teoh, Choi, and Qin]{tnnls3}
Y.~Song, J.~Y.-C. Teoh, K.-S. Choi, and J.~Qin, ``Dynamic loss weighting for multiorgan segmentation in medical images,'' \emph{IEEE Transactions on Neural Networks and Learning Systems}, 2023.

\bibitem[Song et~al.(2023{\natexlab{b}})Song, Teoh, Choi, and Qin]{tnnls4}
------, ``Dynamic loss weighting for multiorgan segmentation in medical images,'' \emph{IEEE Transactions on Neural Networks and Learning Systems}, 2023.

\bibitem[Wang et~al.(2023)Wang, Tang, Xiao, Zhou, Fang, and Yang]{tnnls5}
J.~Wang, Y.~Tang, Y.~Xiao, J.~T. Zhou, Z.~Fang, and F.~Yang, ``Grenet: Gradually recurrent network with curriculum learning for 2-d medical image segmentation,'' \emph{IEEE Transactions on Neural Networks and Learning Systems}, 2023.

\bibitem[Xu et~al.(2023{\natexlab{a}})Xu, Tian, Liu, Wang, Yuan, Gu, Chen, Lukasiewicz, and Leung]{tnnls6}
Z.~Xu, B.~Tian, S.~Liu, X.~Wang, D.~Yuan, J.~Gu, J.~Chen, T.~Lukasiewicz, and V.~C. Leung, ``Collaborative attention guided multi-scale feature fusion network for medical image segmentation,'' \emph{IEEE Transactions on Network Science and Engineering}, 2023.

\bibitem[Li et~al.(2018{\natexlab{b}})Li, Chen, Qi, Dou, Fu, and Heng]{li2018h}
X.~Li, H.~Chen, X.~Qi, Q.~Dou, C.-W. Fu, and P.-A. Heng, ``H-denseunet: Hybrid densely connected unet for liver and tumor segmentation from ct volumes,'' \emph{IEEE Transactions on Medical Imaging}, vol.~37, no.~12, pp. 2663--2674, 2018.

\bibitem[Li et~al.(2022)Li, Cai, Gao, Li, and Hu]{li2021more}
Y.~Li, W.~Cai, Y.~Gao, C.~Li, and X.~Hu, ``More than encoder: Introducing transformer decoder to upsample,'' in \emph{BIBM}.\hskip 1em plus 0.5em minus 0.4em\relax IEEE, 2022, pp. 1597--1602.

\bibitem[Xiao et~al.(2018)Xiao, Lian, Luo, and Li]{xiao2018weighted}
X.~Xiao, S.~Lian, Z.~Luo, and S.~Li, ``Weighted res-unet for high-quality retina vessel segmentation,'' in \emph{ITME}.\hskip 1em plus 0.5em minus 0.4em\relax IEEE, 2018, pp. 327--331.

\bibitem[Lee et~al.(2017)Lee, Zung, Li, Jain, and Seung]{lee2017superhuman}
K.~Lee, J.~Zung, P.~Li, V.~Jain, and H.~S. Seung, ``Superhuman accuracy on the snemi3d connectomics challenge,'' \emph{arXiv preprint arXiv:1706.00120}, 2017.

\bibitem[Zhu et~al.(2020)Zhu, Zhao, Li, Roth, Xu, and Xu]{zhu2020lamp}
W.~Zhu, C.~Zhao, W.~Li, H.~Roth, Z.~Xu, and D.~Xu, ``Lamp: Large deep nets with automated model parallelism for image segmentation,'' in \emph{MICCAI}.\hskip 1em plus 0.5em minus 0.4em\relax Springer, 2020, pp. 374--384.

\bibitem[Kim et~al.(2019)Kim, Kim, Lim, Baek, Kim, Cho, Yoon, and Kim]{kim2019scalable}
S.~Kim, I.~Kim, S.~Lim, W.~Baek, C.~Kim, H.~Cho, B.~Yoon, and T.~Kim, ``Scalable neural architecture search for 3d medical image segmentation,'' in \emph{MICCAI}.\hskip 1em plus 0.5em minus 0.4em\relax Springer, 2019, pp. 220--228.

\bibitem[Zhang et~al.(2021)Zhang, Liu, and Hu]{zhang2021transfuse}
Y.~Zhang, H.~Liu, and Q.~Hu, ``Transfuse: Fusing transformers and cnns for medical image segmentation,'' in \emph{MICCAI}.\hskip 1em plus 0.5em minus 0.4em\relax Springer, 2021, pp. 14--24.

\bibitem[Lin et~al.(2022)Lin, Chen, Xu, Zhang, Lu, and Zhang]{lin2022ds}
A.~Lin, B.~Chen, J.~Xu, Z.~Zhang, G.~Lu, and D.~Zhang, ``Ds-transunet: Dual swin transformer u-net for medical image segmentation,'' \emph{IEEE Transactions on Instrumentation and Measurement}, 2022.

\bibitem[You et~al.(2022)You, Zhao, Liu, Dong, Chinchali, Topcu, Staib, and Duncan]{youclass}
C.~You, R.~Zhao, F.~Liu, S.~Dong, S.~Chinchali, U.~Topcu, L.~Staib, and J.~Duncan, ``Class-aware adversarial transformers for medical image segmentation,'' \emph{Neurips}, vol.~35, pp. 29\,582--29\,596, 2022.

\bibitem[Ji et~al.(2021)Ji, Zhang, Wang, Li, Wu, Zhang, and Luo]{ji2021multi}
Y.~Ji, R.~Zhang, H.~Wang, Z.~Li, L.~Wu, S.~Zhang, and P.~Luo, ``Multi-compound transformer for accurate biomedical image segmentation,'' in \emph{MICCAI}.\hskip 1em plus 0.5em minus 0.4em\relax Springer, 2021, pp. 326--336.

\bibitem[Huang et~al.(2022)Huang, Li, Xiao, Shen, and Xu]{huang2022rtnet}
S.~Huang, J.~Li, Y.~Xiao, N.~Shen, and T.~Xu, ``Rtnet: Relation transformer network for diabetic retinopathy multi-lesion segmentation,'' \emph{IEEE Transactions on Medical Imaging}, 2022.

\bibitem[Wang et~al.(2021{\natexlab{b}})Wang, Chen, Ding, Yu, Zha, and Li]{wang2021transbts}
W.~Wang, C.~Chen, M.~Ding, H.~Yu, S.~Zha, and J.~Li, ``Transbts: Multimodal brain tumor segmentation using transformer,'' in \emph{MICCAI}.\hskip 1em plus 0.5em minus 0.4em\relax Springer, 2021, pp. 109--119.

\bibitem[Karimi et~al.(2021)Karimi, Vasylechko, and Gholipour]{karimi2021convolution}
D.~Karimi, S.~D. Vasylechko, and A.~Gholipour, ``Convolution-free medical image segmentation using transformers,'' in \emph{MICCAI}.\hskip 1em plus 0.5em minus 0.4em\relax Springer, 2021, pp. 78--88.

\bibitem[Wang et~al.(2021{\natexlab{c}})Wang, Wei, Wang, Zhou, Zhu, and Qin]{wang2021boundary}
J.~Wang, L.~Wei, L.~Wang, Q.~Zhou, L.~Zhu, and J.~Qin, ``Boundary-aware transformers for skin lesion segmentation,'' in \emph{MICCAI}.\hskip 1em plus 0.5em minus 0.4em\relax Springer, 2021, pp. 206--216.

\bibitem[Yang et~al.(2021)Yang, Myronenko, Wang, Xu, Roth, and Xu]{yang2021t}
D.~Yang, A.~Myronenko, X.~Wang, Z.~Xu, H.~R. Roth, and D.~Xu, ``T-automl: Automated machine learning for lesion segmentation using transformers in 3d medical imaging,'' in \emph{ICCV}, 2021, pp. 3962--3974.

\bibitem[Yun et~al.(2021)Yun, Wang, Chen, Wang, Shen, and Li]{yun2021spectr}
B.~Yun, Y.~Wang, J.~Chen, H.~Wang, W.~Shen, and Q.~Li, ``{SpecTr}: Spectral transformer for hyperspectral pathology image segmentation,'' \emph{arXiv preprint arXiv:2103.03604}, 2021.

\bibitem[Hatamizadeh et~al.(2022{\natexlab{b}})Hatamizadeh, Nath, Tang, Yang, Roth, and Xu]{hatamizadeh2022swin}
A.~Hatamizadeh, V.~Nath, Y.~Tang, D.~Yang, H.~R. Roth, and D.~Xu, ``Swin unetr: Swin transformers for semantic segmentation of brain tumors in mri images,'' in \emph{MICCAI Brainlesion Workshop}.\hskip 1em plus 0.5em minus 0.4em\relax Springer, 2022, pp. 272--284.

\bibitem[Jiang et~al.(2022)Jiang, Zhang, Lin, Dong, Cheng, and Liang]{Jiang2022SwinBTSAM}
Y.~Jiang, Y.~Zhang, X.~Lin, J.~Dong, T.~Cheng, and J.~Liang, ``Swinbts: A method for 3d multimodal brain tumor segmentation using swin transformer,'' \emph{Brain Sciences}, vol.~12, 2022.

\bibitem[Liu et~al.(2022{\natexlab{a}})Liu, Tian, Xu, Yang, Pan, Yan, and Wang]{liu2022phtrans}
W.~Liu, T.~Tian, W.~Xu, H.~Yang, X.~Pan, S.~Yan, and L.~Wang, ``Phtrans: Parallelly aggregating global and local representations for medical image segmentation,'' in \emph{MICCAI}.\hskip 1em plus 0.5em minus 0.4em\relax Springer, 2022, pp. 235--244.

\bibitem[Wang et~al.(2022)Wang, Wang, Luo, Zhou, Zhou, Wang, Li, and Jin]{wang2021scaled}
P.~Wang, X.~Wang, H.~Luo, J.~Zhou, Z.~Zhou, F.~Wang, H.~Li, and R.~Jin, ``Scaled relu matters for training vision transformers,'' in \emph{AAAI}, vol.~36, no.~3, 2022, pp. 2495--2503.

\bibitem[Touvron et~al.(2021)Touvron, Cord, Douze, Massa, Sablayrolles, and J{\'e}gou]{touvron2021training}
H.~Touvron, M.~Cord, M.~Douze, F.~Massa, A.~Sablayrolles, and H.~J{\'e}gou, ``Training data-efficient image transformers \& distillation through attention,'' in \emph{ICML}.\hskip 1em plus 0.5em minus 0.4em\relax PMLR, 2021, pp. 10\,347--10\,357.

\bibitem[Ren et~al.(2022{\natexlab{a}})Ren, Zhou, He, Feng, and Wang]{ren2022shunted}
S.~Ren, D.~Zhou, S.~He, J.~Feng, and X.~Wang, ``Shunted self-attention via multi-scale token aggregation,'' in \emph{CVPR}, 2022, pp. 10\,853--10\,862.

\bibitem[Ren et~al.(2022{\natexlab{b}})Ren, Zheng, yu~Zhao, Xu, and Li]{j1}
J.~Ren, Q.~Zheng, Y.~yu~Zhao, X.~Xu, and C.~Li, ``Dlformer: Discrete latent transformer for video inpainting. 2022 ieee,'' in \emph{CVPR}, 2022, pp. 3501--3510.

\bibitem[Ren et~al.(2024)Ren, Chen, Ye, Wu, and Zhu]{j2}
J.~Ren, H.~Chen, T.~Ye, H.~Wu, and L.~Zhu, ``Triplane-smoothed video dehazing with clip-enhanced generalization,'' \emph{International Journal of Computer Vision}, pp. 1--14, 2024.

\bibitem[Ren et~al.(2021)Ren, Liu, Liu, Chen, Han, and He]{r2}
S.~Ren, W.~Liu, Y.~Liu, H.~Chen, G.~Han, and S.~He, ``Reciprocal transformations for unsupervised video object segmentation,'' in \emph{CVPR}, 2021, pp. 15\,455--15\,464.

\bibitem[Ren et~al.(2023)Ren, Yang, Liu, and Wang]{r3}
S.~Ren, X.~Yang, S.~Liu, and X.~Wang, ``Sg-former: Self-guided transformer with evolving token reallocation,'' in \emph{ICCV}, 2023, pp. 6003--6014.

\bibitem[Lee et~al.(2015)Lee, Xie, Gallagher, Zhang, and Tu]{lee2015deeply}
C.-Y. Lee, S.~Xie, P.~Gallagher, Z.~Zhang, and Z.~Tu, ``Deeply-supervised nets,'' in \emph{Artificial intelligence and statistics}.\hskip 1em plus 0.5em minus 0.4em\relax PMLR, 2015, pp. 562--570.

\bibitem[Liu et~al.(2022{\natexlab{b}})Liu, Ning, Cao, Wei, Zhang, Lin, and Hu]{liu2021video}
Z.~Liu, J.~Ning, Y.~Cao, Y.~Wei, Z.~Zhang, S.~Lin, and H.~Hu, ``Video swin transformer,'' in \emph{CVPR}, 2022, pp. 3202--3211.

\bibitem[Yao et~al.(2022)Yao, Hu, Li, Zhai, and Zhang]{chang2021transclaw}
C.~Yao, M.~Hu, Q.~Li, G.~Zhai, and X.-P. Zhang, ``Transclaw u-net: Claw u-net with transformers for medical image segmentation,'' in \emph{ICICSP}, 2022, pp. 280--284.

\bibitem[Xu et~al.(2023{\natexlab{b}})Xu, Zhang, He, and Wu]{xu2021levit}
G.~Xu, X.~Zhang, X.~He, and X.~Wu, ``Levit-unet: Make faster encoders with transformer for medical image segmentation,'' in \emph{PRCV}.\hskip 1em plus 0.5em minus 0.4em\relax Springer, 2023, pp. 42--53.

\bibitem[Huang et~al.(2021)Huang, Deng, Li, and Yuan]{huang2021missformer}
X.~Huang, Z.~Deng, D.~Li, and X.~Yuan, ``{MISSFormer}: An effective medical image segmentation transformer,'' \emph{arXiv preprint arXiv:2109.07162}, 2021.

\bibitem[Milletari et~al.(2016)Milletari, Navab, and Ahmadi]{milletari2016v}
F.~Milletari, N.~Navab, and S.-A. Ahmadi, ``V-net: Fully convolutional neural networks for volumetric medical image segmentation,'' in \emph{3DV}.\hskip 1em plus 0.5em minus 0.4em\relax IEEE, 2016, pp. 565--571.

\bibitem[Kobayashi et~al.(2002)]{ce}
K.~Kobayashi \emph{et~al.}, \emph{Mathematics of information and coding}.\hskip 1em plus 0.5em minus 0.4em\relax American Mathematical Soc., 2002, vol. 203.

\bibitem[Sudre et~al.(2017)Sudre, Li, Vercauteren, Ourselin, and Jorge~Cardoso]{dice}
C.~H. Sudre, W.~Li, T.~Vercauteren, S.~Ourselin, and M.~Jorge~Cardoso, ``Generalised dice overlap as a deep learning loss function for highly unbalanced segmentations,'' in \emph{Deep learning in medical image analysis and multimodal learning for clinical decision support}.\hskip 1em plus 0.5em minus 0.4em\relax Springer, 2017, pp. 240--248.

\bibitem[Bakas et~al.(2018)Bakas, Reyes, Jakab, Bauer, Rempfler, Crimi, Shinohara, Berger, Ha, Rozycki, et~al.]{bakas2018identifying}
S.~Bakas, M.~Reyes, A.~Jakab, S.~Bauer, M.~Rempfler, A.~Crimi, R.~T. Shinohara, C.~Berger, S.~M. Ha, M.~Rozycki \emph{et~al.}, ``Identifying the best machine learning algorithms for brain tumor segmentation, progression assessment, and overall survival prediction in the brats challenge,'' \emph{arXiv preprint arXiv:1811.02629}, 2018.

\bibitem[Bernard et~al.(2018)Bernard, Lalande, Zotti, Cervenansky, Yang, Heng, Cetin, Lekadir, Camara, Ballester, et~al.]{bernard2018deep}
O.~Bernard, A.~Lalande, C.~Zotti, F.~Cervenansky, X.~Yang, P.-A. Heng, I.~Cetin, K.~Lekadir, O.~Camara, M.~A.~G. Ballester \emph{et~al.}, ``Deep learning techniques for automatic mri cardiac multi-structures segmentation and diagnosis: Is the problem solved?'' \emph{IEEE Transactions on Medical Imaging}, vol.~37, no.~11, pp. 2514--2525, 2018.

\bibitem[Zheng et~al.(2021)Zheng, Lu, Zhao, Zhu, Luo, Wang, Fu, Feng, Xiang, Torr, et~al.]{zheng2021rethinking}
S.~Zheng, J.~Lu, H.~Zhao, X.~Zhu, Z.~Luo, Y.~Wang, Y.~Fu, J.~Feng, T.~Xiang, P.~H. Torr \emph{et~al.}, ``Rethinking semantic segmentation from a sequence-to-sequence perspective with transformers,'' in \emph{CVPR}, 2021, pp. 6881--6890.

\bibitem[Gu and Dao(2023)]{mamba1}
A.~Gu and T.~Dao, ``Mamba: Linear-time sequence modeling with selective state spaces,'' \emph{arXiv preprint arXiv:2312.00752}, 2023.

\bibitem[Dao and Gu(2024)]{mamba2}
T.~Dao and A.~Gu, ``Transformers are {SSM}s: Generalized models and efficient algorithms through structured state space duality,'' in \emph{ICML}, ser. Proceedings of Machine Learning Research, R.~Salakhutdinov, Z.~Kolter, K.~Heller, A.~Weller, N.~Oliver, J.~Scarlett, and F.~Berkenkamp, Eds., vol. 235.\hskip 1em plus 0.5em minus 0.4em\relax PMLR, 21--27 Jul 2024, pp. 10\,041--10\,071.

\bibitem[Kirillov et~al.(2023)Kirillov, Mintun, Ravi, Mao, Rolland, Gustafson, Xiao, Whitehead, Berg, Lo, et~al.]{sam}
A.~Kirillov, E.~Mintun, N.~Ravi, H.~Mao, C.~Rolland, L.~Gustafson, T.~Xiao, S.~Whitehead, A.~C. Berg, W.-Y. Lo \emph{et~al.}, ``Segment anything,'' in \emph{ICCV}, 2023, pp. 4015--4026.

\end{thebibliography}

\begin{IEEEbiography}[{\includegraphics[width=1in,height=1.25in,clip,keepaspectratio]{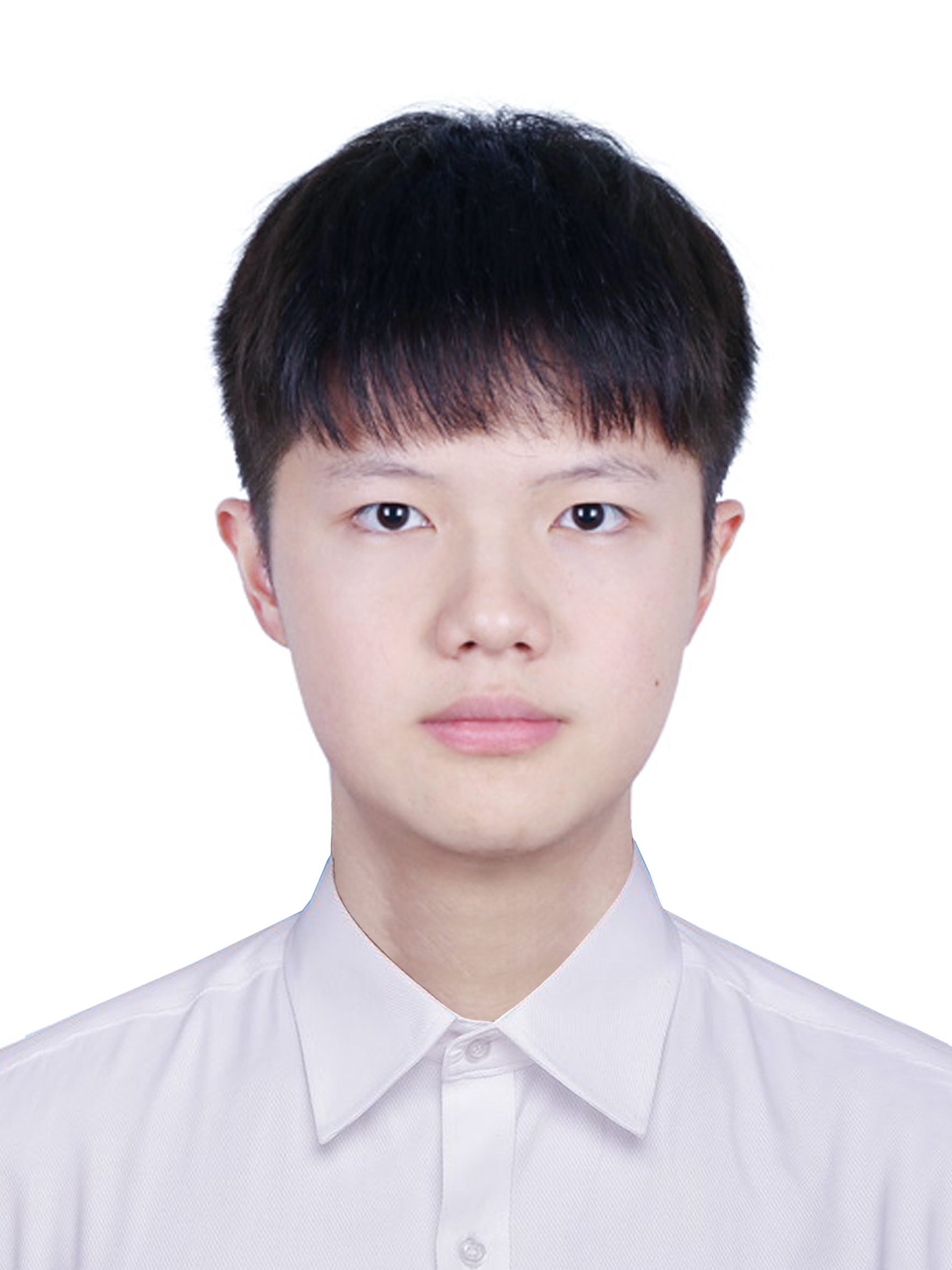}}]{Sucheng Ren} is a research assistant in the department of Electronic and Computer Engineering, The Hong Kong University of Science and Technology. He received the M.Sc. degree and B. Eng. degree from South China University of Technology. His research interests include computer vision, image processing, and deep learning.
\end{IEEEbiography}

\begin{IEEEbiography}[{\includegraphics[width=1in,height=1.25in,clip,keepaspectratio]{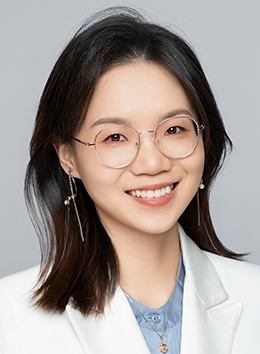}}]{Xiaomeng Li} received her PhD in the Department of Computer Science and Technology at the Chinese University of Hong Kong. She is currently an Assistant Professor at the Hong Kong University of Science and Technology in Hong Kong SAR, China. Her research focuses on developing advanced AI methods to enhance healthcare, particularly in medical image analysis, including image segmentation and classification.

\end{IEEEbiography}

\end{document}